\documentclass[11pt]{article}
\usepackage[utf8]{inputenc}
\usepackage[T1]{fontenc}
\usepackage{graphicx}
\usepackage{comment}
\usepackage{rotating}
\usepackage[normalem]{ulem}
\usepackage{amsmath}
\usepackage{amssymb}
\usepackage{capt-of}
\usepackage{subcaption} 
\usepackage{hyperref}

\usepackage{lrec2026}
\usepackage{times}
\usepackage{soul}
\usepackage{enumitem}

\title{Align and Shine: Building High-Quality Sentence-Aligned Corpora for Multilingual Text Simplification}
\usepackage{natbib}
\date{}
\name{Kenji Hilasaca, Nouran Khallaf, Serge Sharoff} 

\address{Centre for Translation, Localisation and Interpreting Studies \\
School of Languages, Cultures and Societies\\
         University of Leeds, UK \\
         pjbd103@leeds.ac.uk, n.khallaf@leeds.ac.uk, s.sharoff@leeds.ac.uk}

\abstract{Text simplification plays a crucial role in improving the accessibility and comprehensibility of written information for diverse audiences, including language learners and readers with limited literacy. Despite its importance, large-scale, high-quality datasets for training and evaluating text simplification models remain scarce for languages other than English. This paper reports an experimental study on the collection and processing of crowd-sourced simplification data from comparable corpora to construct a corpus suitable for both training and testing text simplification systems across multiple languages (Catalan, English, French, Italian and Spanish). We report mechanisms for sentence-level alignment from document-level data. The resulting dataset of the aligned sentence pairs is publicly available.
\\
\newline }

\begin{document}

\maketitleabstract
\setlist{itemsep=5pt,parsep=0pt}
\section{Introduction}
\label{sec:org12c9bbf}

Automatic text simplification plays a crucial role in improving the accessibility and comprehensibility of written information for diverse audiences, including language learners and readers with limited literacy \citep{saggion2017simplification}. Data needed for training automatic text simplification tools are based on aligned sentences. This alignment at the sentence level, rather than at the document level, is essential for supervised learning approaches, as it enables models to learn specific simplification operations and their contextual application. A sentence-aligned corpus is also essential for evaluation of zero-shot approaches.


Despite its importance, large-scale, high-quality publicly available datasets for training and evaluating text simplification models remain scarce for languages other than English, which has such datasets as ASSET \citep{alvamanchego20asset} and Wikipedia-derived ParallelSEW \citep{coster11parallelsew}. This paper reports an experimental study on collecting and processing of crowd-sourced simplification data to construct a corpus suitable for both training and testing text simplification systems across multiple languages. We report mechanisms for sentence-level alignment from document-level data. The resulting dataset, together with the aligned sentence pairs, is publicly available.\footnote{\url{https://github.com/kenjihilasak/Align-and-Shine}} This is the first large open-source corpus for text simplification in Catalan and Spanish. The parallel corpus is also consistent across the languages with respect to its genre, which will help with cross-lingual evaluation of text simplification models.

\section{Related studies}
\label{sec:org5f3d13d}
Early work on machine translation highlighted both the value and the limitations of domain-specific resources such as the European Parliament corpus \cite{koehn05} and the United Nations corpus \cite{ziemski16}. The limitations on the amount and diversity of texts motivated large-scale mining from comparable corpora, a line of research that ultimately contributed to the pre-training data pipelines for Large Language Models \cite{sharoff23bucc}. Within this paradigm, Wikipedia articles connected via iWiki links proved a particularly productive source for extracting translation pairs \cite{adafre06,schwenk19wiki}, owing to their broad topical coverage and cross-lingual parallelism. The sentence alignment methods have also evolved considerably, from early character-based approaches \cite{gale93} through more advanced statistical models \cite{varga07} to contemporary neural architectures \cite{jiang2020neural}.

Similar efforts extracted complex-simple pairs from web corpora \cite{brunato16} or machine-translated Wikipedia alignments to construct multilingual resources \cite{cardon2020french}. While these were consolidated into a single corpus \cite{ryan23tsnonEnglish}, the result suffers from noise and inconsistent annotation due to domain mismatch.

Our contribution addresses this gap directly: we conduct alignment experiments on a novel, reliable dataset drawn from a single source domain across multiple languages, using modern sentence-alignment tools. Furthermore, while recent advancements have introduced aligners, such as CATS \citep{stajner2018cats}, as well as robust multilingual models like Bertalign \citep{niklaus2026}, our current focus is on evaluating the impact of distinct semantic representation paradigms within the hybrid SentAlign framework, leaving the comparison against these newer aligners for future work.


\begin{table*}[t]
\caption{Initial Wikipedia / Vikidia corpus. \#Docs is the document count (paired Wikipedia--Vikidia articles by topic). \#Words and \#Sentences are totals across all documents in each subset. IQR gives the inter-quartile (25\% to 75\%) range of sentence lengths in words in each subset. \label{tabCorpus}}
\centering
\small
\begin{tabular}{lr|rrr|rrr} 
& & \multicolumn{3}{c|}{\textbf{Wikipedia}}& \multicolumn{3}{c}{\textbf{Vikidia}}\\
\textbf{Language} & \textbf{\#Docs} & \textbf{\#Words} & \textbf{\#Sent's} & \textbf{IQR} & \textbf{\#Words} & \textbf{\#Sent's} & \textbf{IQR}\\
\hline
Catalan & 179 & 396,277 & 16,813 & (15, 29) & 19,394 & 1,000 & (13, 23)\\
English & 2,585 & 8,281,625 & 340,924 & (16, 29) & 424,306 & 22,462 & (13, 22)\\
Spanish & 3,875 & 7,946,169 & 301,241 & (16, 33) & 607,990 & 27,825 & (14, 26)\\
French & 33,438 & 46,618,143 & 1,945,046 & (15, 29) & 6,643,567 & 320,372 & (14, 25)\\
Italian & 3,902 & 6,790,163 & 263,271 & (16, 32) & 537,723 & 25,202 & (14, 26)\\
\end{tabular}
\end{table*}

\section{Methodology}
\label{sec:methodology}

Our study proposes a two-phase methodology for identifying parallel sentences in document-aligned simplification corpora. First, we evaluate traditional surface-level baselines against modern semantic embedding methods using a manually annotated gold standard. This evaluation phase allows us to identify the best-performing embedding model and tune the optimal cosine similarity threshold ($\tau$) to filter out noise and verbatim copies for each language. In the second phase, we apply these optimal, language-specific configurations to a large Web-derived comparable corpus to extract a large-scale dataset suitable for text simplification. To accurately capture text simplification operations, which inherently involve compression, summarization or explanation, our alignment pipeline utilizes an asymmetric search strategy that supports flexible $N$-to-$M$ mappings.

\subsection{Baseline Approaches}
\label{sec:baselines}

We evaluate two established language-independent baselines to quantify the benefits of neural embeddings.

The \textbf{Gale and Church} algorithm \citep{gale93} assumes correlated sentence lengths in parallel texts. While this assumption is often violated in text simplification due to sentence splitting or compression, it is evaluated here because it forms the heuristic pre-selection stage of the primary framework (SentAlign). Testing it independently serves as a necessary ablation baseline to demonstrate the value added by the subsequent neural semantic anchoring stage.

\textbf{Hunalign} \citep{varga07} augments length heuristics with a dictionary built from parallel texts. While effective for translation alignment, its reliance on lexical overlap limits performance in text simplification tasks.

\subsection{Experiments with SentAlign}
\label{sec:sentalign_framework}

To overcome the limitations of length-based methods, we use \textbf{SentAlign} \citep{steingrimsson23sentalign}, a hybrid alignment algorithm combining neural embeddings with a three-stage pipeline:

\begin{enumerate}
    \item \textbf{Heuristic Pre-selection:} SentAlign reduces the search space by generating candidate alignments using the \citet{gale93} algorithm, which relies on character length ratios. 

    \item \textbf{Semantic Anchoring:} Candidates are validated using the chosen embedding model (Section \ref{sec:embeddings}). Those exceeding a high-confidence cosine similarity threshold become \textit{anchors}. This establishes fixed points in the document map that partition the text into smaller segments.

    \item \textbf{Global Optimization:} The algorithm aligns segments between anchors using Dijkstra's shortest path algorithm, with costs derived from the cosine similarity matrix. We configure this stage to prioritize the simplified document as the reference, enabling an asymmetric search that retrieves the closest semantic equivalent(s) in the complex document for each simple sentence. This easily accommodates flexible $N$-to-$M$ mappings (e.g., $1$-to-$2$, $1$-to-$0$, or $2$-to-$1$), successfully capturing simplification operations such as splitting, deletion, and summarization.
\end{enumerate}

\subsection{Comparison of Semantic Representations}
\label{sec:embeddings}

Our experiment focuses on evaluating the impact of different semantic representations during the \textit{Semantic Anchoring} and \textit{Global Optimization} stages of SentAlign. We compare three multilingual encoders:

\begin{itemize}
    \item \textbf{LaBSE} \citep{feng22labse}: A translation-optimized model, traditionally strong in cross-lingual alignment. We assess its transferability to monolingual text simplification tasks.

    \item \textbf{BGE-M3} \citep{xiao24bge}: Designed for Retrieval-Augmented Generation (RAG) and semantic search, BGE-M3 handles multi-granularity inputs. We hypothesize it may excel in simplification due to its focus on retrieving the most relevant sentences, and its ability to manage structural and length disparities between complex and simple text.

    \item \textbf{SONAR} \citep{duquenne23sonar}: Developed under the \textit{No Language Left Behind} (NLLB) project, SONAR supports over 200 languages using a distinct architecture. We test whether its massive multilingual capacity improves alignment for lower-resource languages (e.g., Catalan in our case) compared to BERT-based models.
\end{itemize}


\section{Evaluation Setup}
\label{sec:evaluation}

We start with the initial corpus, which has been crawled from Vikidia\footnote{\url{https://www.vikidia.org/}}, a website that maintains Wikipedia-style content aimed at ``children and anyone seeking easy-to-read content''. For each Vikidia document, we added the corresponding Wikipedia article in the same language to form comparable document pairs. Stub articles (with little content at the moment) have been discarded. The total amount of data across all languages is listed in Table \ref{tabCorpus}. The entries have been aligned between the two versions when they had identical headings, hence we have the same document count for each language. However, the Vikidia entries are much shorter, so the sentence count for Vikidia is 10--15 times smaller for our languages apart from French, as Vikidia is much more popular in the French-speaking world. The IQR column shows that the sentences in Vikidia are consistently shorter, but not much shorter than their Wikipedia counterparts.

To assess the performance of the different alignment configurations, we perform an intrinsic evaluation focusing on the accurate retrieval of simplified sentence pairs.

\subsection{Gold Standard Creation}
\label{sec:gold_standard}

To create a reliable ground truth for our intrinsic evaluation, we randomly selected 15 document pairs for each language. The sentence alignment for these documents was performed manually from scratch by one annotator per language.

To ensure high inter-annotator agreement and to capture the nature of text simplification more accurately, the annotators followed our guidelines. Two sentences (or groups of sentences) were considered aligned if they exhibited a clear semantic correspondence, specifically encompassing the following accepted simplification operations:
\begin{itemize}
    \item \textbf{Lexical Simplification (1-to-1):} Substituting complex vocabulary or idioms with accessible equivalents.
    \item \textbf{Syntactic Simplification:} Restructuring complex grammar (e.g., passive to active voice).
    \item \textbf{Sentence Splitting (1-to-N):} Breaking a long complex sentence into multiple shorter ones.
    \item \textbf{Summarization (N-to-1):} Condensing peripheral details from multiple sentences into a single core sentence.
    \item \textbf{Block Summarization (N-to-N):} Reorganizing complex paragraphs into a different number of simplified sentences while maintaining semantic equivalence.
    \item \textbf{Deletion:} Omitting overly technical or tangential sentences entirely (deliberately left unaligned).
\end{itemize}


\subsection{Evaluation Metrics and Thresholding}
\label{sec:eval_metrics}

For each alignment tool and embedding variant, we compute the similarities across the entire paired document. The performance is then measured using standard retrieval metrics: \textbf{Precision}, \textbf{Recall}, and \textbf{F$_1$-score}. These metrics are calculated by comparing the system's output against our manually annotated Gold Standard using the official SentAlign evaluation script\footnote{\url{https://github.com/steinst/SentAlign/blob/master/evaluation/evaluate.py}}.

Because the raw output includes pairs with varying degrees of semantic overlap, we implement a filtering mechanism to isolate actual simplifications:
\begin{itemize}
    \item \textbf{Lower-bound Threshold ($\tau$):} For each language and model configuration, we establish a specific threshold ($\tau$) tuned via grid search to maximize the F$_1$-score on the test sample.
    \item \textbf{Upper-bound Threshold (0.95):} To explicitly capture text where simplification operations have occurred (e.g., paraphrasing, lexical substitution, or splitting), we discard sentence pairs with a cosine similarity score $> 0.95$. This removes exact or near-exact copies (with minor variations in punctuation) that provide no learning signal for text simplification models, and tend to plague such alignments as WikiLarge \cite{cardon2020french}. The upper bound threshold was found to be acceptable for all languages and for all embedding frameworks.
\end{itemize}

Only the sentence pairs whose scores fall within this $[\tau, 0.95]$ range are considered acceptable alignments and passed to the evaluation script. We consolidate the results using these optimal thresholds, yielding two evaluation settings based on how forgiving the scoring is when the algorithm makes a "partial match" compared to the Gold Standard. \textbf{Strict Evaluation} requires an exact group match (e.g., if the Gold Standard establishes a 2-to-1 summarization like [1, 2] $\rightarrow$ [3], the algorithm only gets credit if it predicts exactly [1, 2] $\rightarrow$ [3]). Conversely, \textbf{Lax Evaluation} allows for partial overlap, rewarding the algorithm if it successfully finds a valid, albeit incomplete, semantic connection (e.g., predicting [1] $\rightarrow$ [3] when the true label is [1, 2] $\rightarrow$ [3]).

\setlength{\tabcolsep}{3pt}
\begin{table*}[!t]
\caption{Comprehensive alignment results comparing raw outputs (None) and optimized threshold outputs ($\tau$). \textbf{Strict} requires exact matches, while \textbf{Lax} rewards partial semantic overlaps. Best F$_1$ scores per language are highlighted in bold. \label{tab:main_results}}
\centering
\small
\setlength{\tabcolsep}{4pt}
\begin{tabular}{ll c | rrr | rrr}
\hline
& & & \multicolumn{3}{c|}{\textbf{Strict Evaluation}} & \multicolumn{3}{c}{\textbf{Lax Evaluation}} \\
\textbf{Language} & \textbf{Algorithm} & \textbf{Threshold ($\tau$)} & \textbf{P} & \textbf{R} & \textbf{F$_1$} & \textbf{P} & \textbf{R} & \textbf{F$_1$} \\
\hline
\textbf{Catalan} & Gale-Church & None & 0.000 & 0.000 & 0.000 & 0.000 & 0.000 & 0.000 \\ 
(ca)             & Hunalign    & None & 0.000 & 0.000 & 0.000 & 0.000 & 0.000 & 0.000 \\ 
                 & LaBSE       & None & 0.215 & 0.950 & 0.351 & 0.223 & 0.952 & 0.361 \\
                 & \textbf{LaBSE} & \textbf{0.61} & \textbf{0.526} & 0.833 & \textbf{0.645} & \textbf{0.547} & 0.839 & \textbf{0.662} \\
                 & BGE         & None & 0.140 & 0.667 & 0.232 & 0.182 & 0.722 & 0.291 \\
                 & BGE         & 0.58 & 0.276 & 0.617 & 0.381 & 0.366 & 0.681 & 0.476 \\
                 & SONAR       & None & 0.160 & 0.433 & 0.233 & 0.215 & 0.507 & 0.302 \\
                 & SONAR       & 0.56 & 0.302 & 0.267 & 0.283 & 0.396 & 0.323 & 0.356 \\
\hline
\textbf{English} & Gale-Church & None & 0.000 & 0.000 & 0.000 & 0.000 & 0.000 & 0.000 \\
(en)             & Hunalign    & None & 0.020 & 0.090 & 0.030 & 0.020 & 0.090 & 0.030 \\ 
                 & LaBSE       & None & 0.218 & 0.672 & 0.330 & 0.277 & 0.722 & 0.400 \\
                 & LaBSE       & 0.67 & 0.570 & 0.414 & 0.480 & 0.688 & 0.460 & 0.552 \\
                 & BGE         & None & 0.111 & 0.711 & 0.192 & 0.139 & 0.755 & 0.235 \\
                 & \textbf{BGE}& \textbf{0.73} & 0.562 & 0.492 & \textbf{0.525} & \textbf{0.634} & 0.522 & \textbf{0.573} \\
                 & SONAR       & None & 0.188 & 0.367 & 0.249 & 0.264 & 0.449 & 0.332 \\
                 & SONAR       & 0.53 & 0.325 & 0.289 & 0.306 & 0.421 & 0.345 & 0.379 \\
\hline
\textbf{Spanish} & Gale-Church & None & 0.000 & 0.000 & 0.000 & 0.000 & 0.000 & 0.000 \\
(es)             & Hunalign    & None & 0.070 & 0.270 & 0.110 & 0.070 & 0.270 & 0.110 \\
                 & LaBSE       & None & 0.135 & 0.694 & 0.227 & 0.171 & 0.742 & 0.279 \\
                 & \textbf{LaBSE} & \textbf{0.67} & \textbf{0.418} & 0.426 & \textbf{0.422} & \textbf{0.509} & 0.475 & \textbf{0.491} \\
                 & BGE         & None & 0.074 & 0.500 & 0.129 & 0.103 & 0.581 & 0.175 \\
                 & BGE         & 0.71 & 0.330 & 0.343 & 0.336 & 0.464 & 0.423 & 0.443 \\
                 & SONAR       & None & 0.146 & 0.333 & 0.203 & 0.215 & 0.424 & 0.285 \\
                 & SONAR       & 0.58 & 0.253 & 0.222 & 0.236 & 0.263 & 0.229 & 0.245 \\
\hline
\textbf{French}  & Gale-Church & None & 0.000 & 0.000 & 0.000 & 0.000 & 0.000 & 0.000 \\
(fr)             & Hunalign    & None & 0.030 & 0.090 & 0.040 & 0.030 & 0.090 & 0.040 \\
                 & LaBSE       & None & 0.166 & 0.660 & 0.266 & 0.201 & 0.701 & 0.312 \\
                 & \textbf{LaBSE} & \textbf{0.62} & \textbf{0.451} & 0.489 & \textbf{0.469} & \textbf{0.510} & 0.520 & \textbf{0.515} \\
                 & BGE         & None & 0.061 & 0.340 & 0.104 & 0.126 & 0.516 & 0.202 \\
                 & BGE         & 0.70 & 0.292 & 0.223 & 0.253 & 0.569 & 0.360 & 0.441 \\
                 & SONAR       & None & 0.111 & 0.255 & 0.155 & 0.190 & 0.369 & 0.251 \\
                 & SONAR       & 0.54 & 0.190 & 0.202 & 0.196 & 0.240 & 0.242 & 0.241 \\
\hline
\textbf{Italian} & Gale-Church & None & 0.000 & 0.000 & 0.000 & 0.000 & 0.000 & 0.000 \\
(it)             & Hunalign    & None & 0.070 & 0.350 & 0.110 & 0.070 & 0.350 & 0.110 \\
                 & LaBSE       & None & 0.129 & 0.831 & 0.223 & 0.147 & 0.849 & 0.250 \\
                 & \textbf{LaBSE} & \textbf{0.64} & \textbf{0.448} & 0.727 & \textbf{0.554} & \textbf{0.488} & 0.744 & \textbf{0.589} \\
                 & BGE         & None & 0.072 & 0.662 & 0.130 & 0.090 & 0.711 & 0.160 \\
                 & BGE         & 0.78 & 0.494 & 0.494 & 0.494 & 0.597 & 0.541 & 0.568 \\
                 & SONAR       & None & 0.119 & 0.416 & 0.186 & 0.138 & 0.451 & 0.211 \\
                 & SONAR       & 0.55 & 0.213 & 0.351 & 0.265 & 0.236 & 0.375 & 0.290 \\
\hline
\end{tabular}
\end{table*}


\section{Results and Discussion}
\label{sec:results}

\subsection{Quantitative Analysis}
\label{sec:quant_analysis}

\begin{figure*}[!t]
    \centering
    \includegraphics[width=\textwidth]{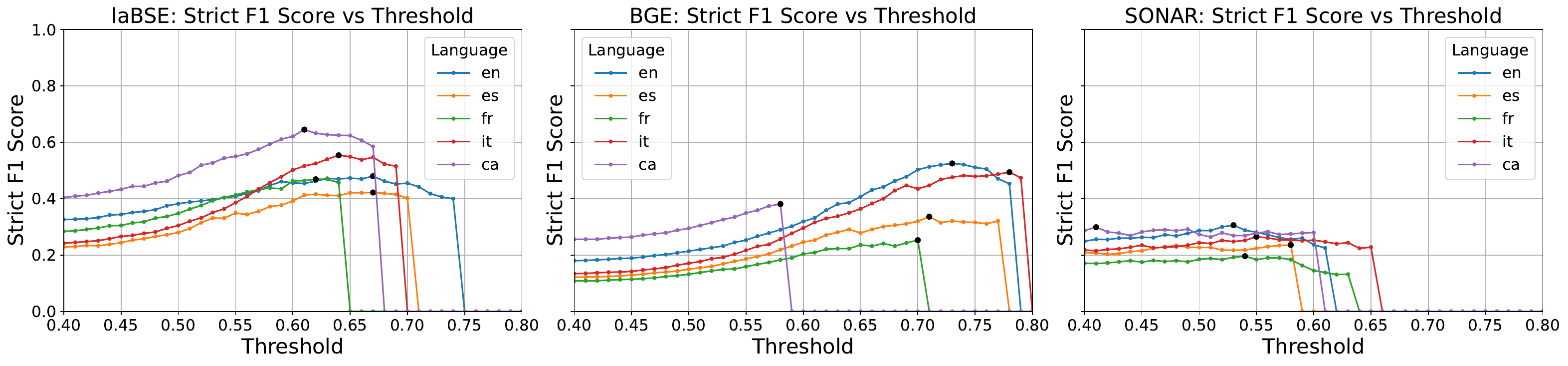} 
    \caption{Plotting Strict F$_1$-scores across cosine similarity thresholds ($\tau$) for the five languages for each embedding method. The black dots indicate the optimal threshold that maximizes the F$_1$-score for each language and for each method. Notice the overall height superiority of LaBSE, the rightward shift of the BGE peaks reflecting its similarity distribution, and the compressed performance of SONAR.}
    \label{fig:threshold_evolution}
\end{figure*}

Table~\ref{tab:main_results} presents the comprehensive alignment performance across all five languages. We report the two surface-feature baselines alongside the SentAlign framework instantiated with three embedding spaces (LaBSE, BGE-M3, and SONAR). For each neural model, we include both raw output (without threshold filtering) and optimised output (using the best threshold $\tau$ per model and language), enabling a direct comparison between the intrinsic structure of each vector space and its tuned performance.

\paragraph{Baseline methods.}
There is a stark contrast between traditional and embedding-based approaches. Gale-Church and Hunalign fail to produce viable alignments in the text simplification setting. Both methods rely on sentence-length correlation and lexical overlap, assumptions that break down when confronted with the heavy paraphrasing, sentence splitting, and compression that characterise Wikipedia-to-Vikidia aligned corpus. Consequently, neither baseline yields F$_1$-scores above chance, confirming that surface-level features are insufficient for this task.

\paragraph{LaBSE.}
Among the neural approaches, \textbf{LaBSE} demonstrates consistent superiority across the Romance languages. In the optimised Strict evaluation, it achieves the highest F$_1$-scores in Catalan (0.645), Spanish (0.422), French (0.469), and Italian (0.554). Its stability across both high- and lower-resource settings suggests that LaBSE's translation-ranking training objective produces a semantic space well suited to detecting monolingual paraphrases, a capability that directly benefits the simplification alignment task.

\paragraph{BGE-M3.}
\textbf{BGE-M3} exhibits a clear domain-specific advantage in English, where it is the only model to surpass LaBSE (Strict F$_1$: 0.525 vs.\ 0.480). This is consistent with its architecture, which is heavily optimised for English-centric retrieval tasks. However, performance degrades markedly in French (Strict F$_1$: 0.253) and Spanish (0.336), pointing to limited cross-lingual stability. The model's strong English performance should therefore be interpreted in light of this imbalance rather than as evidence of general robustness.

\paragraph{SONAR.}
Despite its extensive multilingual pre-training, \textbf{SONAR} trails both LaBSE and BGE across all languages, reaching a maximum Strict F$_1$ of only 0.306 in English. We attribute this to a mismatch between SONAR's training objective and the demands of the present task: while SONAR excels at cross-lingual alignment, its vector space appears to lack the fine-grained monolingual resolution required to reliably distinguish a genuine simplification from a merely topically related sentence within the same language.

\paragraph{Strict vs.~Lax evaluation.}
Across all optimized models, \textbf{Lax F$_1$-scores} consistently exceed their Strict counterparts. LaBSE's English score, for instance, rises from 0.480 (Strict) to 0.552 (Lax). This gap confirms that the SentAlign framework captures partial simplification operations, such as 1-to-$N$ sentence splits, that are penalised under exact-match criteria despite constituting semantically valid alignments. The Lax metric thus provides a more faithful upper-bound estimate of true alignment quality.

\subsection{Impact of Thresholding}
\label{sec:thresholding}

The inclusion of unfiltered metrics in Table~\ref{tab:main_results} highlights the critical role of threshold selection in semantic alignment.  In the absence of filtering, models default to near-exhaustive recall at the cost of precision: LaBSE in Catalan, for example, achieves a raw recall of 0.950 while precision collapses to 0.215, as the algorithm aligns almost every target sentence regardless of semantic relevance. Applying the optimal $\tau$ corrects this imbalance, raising precision to 0.526 and restoring a competitive F$_1$-score.

Figure~\ref{fig:threshold_evolution} plots the Strict F$_1$-score as a function of $\tau$ for all three models and five languages on a unified scale. The shape and position of the curves are informative about the underlying geometry of its vector space:

\begin{description}
   
    \item [LaBSE] produces the highest performance across most languages, with optimal thresholds clustered between $\tau = 0.60$ and $0.67$. Its curves rise gradually before dropping sharply to zero beyond the optimum, a cliff-edge pattern indicating that overly conservative thresholds aggressively discard valid paraphrases. This sensitivity reinforces the necessity of careful per-language tuning.

    \item [BGE-M3] shows optimal thresholds shifted substantially rightward, predominantly between $0.70$ and $0.78$. This reflects systematically higher cosine similarity scores in BGE-M3's vector space, requiring a stricter $\tau$ to separate true matches from background noise. The model also exhibits the greatest cross-lingual variance: English and Italian peak comparably to LaBSE, while French remains markedly depressed, consistent with the quantitative results in Table~\ref{tab:main_results}.

    \item [SONAR] occupies the bottom of the performance range across all languages. Its curves not only peak lower but decay earlier, collapsing before $\tau = 0.65$. This premature decay reflects the model's difficulty in producing high-confidence semantic links for monolingual simplification alignment, corroborating the quantitative findings and suggesting a fundamental mismatch between SONAR's pre-training regime and the demands of this task.
\end{description}

\subsection{Qualitative Analysis and Limitations}
\label{sec:error_analysis}

\begin{table*}[t]
\caption{Aligned Wikipedia / Vikidia corpus. Meaning preservation needs to be close to 1. For the Simplification metrics, $\Delta$ SDepth and $\Delta$ NDense represent the average change (Target - Source) within each alignment block. \label{tabSentences}}
\centering
\small
\begin{tabular}{l|rr|rr|r|rr}
\hline
& \multicolumn{2}{c|}{\textbf{Wikipedia}}& \multicolumn{2}{c|}{\textbf{Vikidia}} & \textbf{Meaning} &  \multicolumn{2}{c}{\textbf{Simplification}} \\
\textbf{Language} & \textbf{\#Words} & \textbf{\#Sent's} & \textbf{\#Words} & \textbf{\#Sent's} & \textbf{BERTScore} & \textbf{$\Delta$ SDepth} & \textbf{$\Delta$ NDense}\\
\hline
Catalan (LaBSE) & 13,658 & 501 & 8,455 & 464 & 0.902 & -0.98 & +4.07\% \\
English (BGE)   & 115,326 & 5,058 & 79,372 & 4,569 & 0.913 & -0.78 & +2.18\% \\
Spanish (LaBSE) & 204,928 & 7,883 & 153,718 & 7,209 & 0.915 & -0.42 & +1.62\% \\
French (LaBSE)  & 2,591,525 & 123,280 & 2,046,642 & 108,301 & 0.901 & -0.23 & +1.20\% \\
Italian (LaBSE) & 260,142 & 11,131 & 185,681 & 9,846 & 0.908 & -0.44 & +1.31\% \\
\hline
\end{tabular}
\end{table*}

To complement our quantitative evaluation, we conducted a manual inspection of the alignment outputs across the five languages to understand the models' behavior and identify typical errors. We categorize the aligned pairs into three distinct regions based on their similarity scores, revealing both the strengths of the pipeline and the inherent challenges of mining parallel data from independently edited wikis.

\paragraph{The Upper Bound: Verbatim copies ($>0.95$).}
A distinct class at the extreme high end of the distribution consists of sentence pairs that are essentially identical in wording. These arise when Vikidia retains a Wikipedia sentence verbatim. Qualitative inspection confirms that these pairs typically involve proper nouns, numerical data, or definitional statements that resist paraphrasing. As noted in Section~\ref{sec:methodology}, pairs scoring above $0.95$ are explicitly excluded from the final corpus because they carry no simplification signal and would teach a generation model to simply copy the source text. 

\paragraph{The Sweet Spot: Genuine simplification.}
In the high-to-mid scoring range, the models successfully capture true simplification operations without semantic drift. In these optimal alignments, the Vikidia segments typically contain shorter sentences, exhibit reduced syntactic complexity, substitute technical vocabulary with more accessible alternatives, and omit parenthetical qualifying clauses, all while keeping the propositional core intact. This confirms that contextual embeddings are well-calibrated for detecting semantic equivalence even when structural changes are drastic (e.g., 1-to-N sentence splits).

\paragraph{The Threshold Limit: Topical overlap vs. Propositional equivalence.}
The most typical errors in our pipeline occur close to the $\tau$ threshold limit. Across all five languages, sentence pairs in this lower-scoring boundary often share a common topic and named entities—creating a "topical illusion"—but diverge in propositional content. Both sentences discuss the same subject, yet they assert different facts or approach the topic from distinct perspectives. This reflects a fundamental property of the Vikidia corpus: a substantial proportion of its sentences were not produced by directly simplifying their Wikipedia counterparts, but were instead written independently by editors. Consequently, the pipeline's most common error is aligning sentences that are topically related but semantically mismatched, which strongly justifies the necessity of our strict, empirically tuned thresholds to filter out this noise.

\subsection{Whole corpus alignment}
With better understanding of the best alignment parameters, we have applied the best models (LaBSE with the respective $\tau$ for all languages except English, where BGE was used) to each document pair of the full corpus from Table~\ref{tabCorpus}. The results are presented in Table~\ref{tabSentences}. On average, ~5\% of the original sentences in the Vikidia corpora find high-confidence equivalents in the respective Wikipedia articles. This strict filtering significantly reduces data volume but guarantees a noise-free corpus, which is crucial in text simplification to prevent model hallucinations and teach genuine simplification rather than loose semantic similarities.

To validate the quality of the final aligned corpus, we perform an automated linguistic assessment focusing on two core dimensions of text simplification: Meaning Preservation and Structural Simplification. For this analysis, we evaluate the optimal alignments generated by our best-performing models per language as established in our previous tests.

\textbf{Meaning Preservation:} We utilize BERTScore \citep{zhang20BERTScore} to measure the semantic equivalence between the complex source and the simplified target. Crucially, to accurately evaluate N-to-M alignments (such as sentence splits or multi-sentence summarizations), we concatenated the sentences within each aligned group before computing the score. This ensures the contextual embedding model evaluates the full semantic unit. As shown in Table \ref{tabSentences}, a high average BERTScore F$_1$ across all languages (consistently $>0.90$) confirms that the SentAlign pipeline, when constrained by selecting optimal thresholds ($\tau$) and upper bound filter ($<0.95$), successfully extracts aligned pairs that are very similar in their meaning.

\textbf{Structural Simplification:} To verify that the target sentences are genuinely simpler, we parsed the sentences using SpaCy to compute two syntactic metrics: Maximum Tree Depth and Noun Phrase (NP) Density. For alignments containing multiple sentences, NP Density was calculated by aggregating the total number of NPs divided by the total number of tokens across the aligned sentence group, while Tree Depth was taken as the maximum depth among the constituent sentences within that specific alignment.

The results in Table \ref{tabSentences} reveal the structural mechanisms of the simplifications. The negative values in $\Delta$ Max Tree Depth confirm that the Vikidia target texts consistently employ flatter, less complex grammatical structures. Interestingly, $\Delta$ NP Density exhibits a slight increase across all languages (e.g., $+2.18\%$ in English). This is a well-documented artifact of text distillation: as peripheral words (adverbs, complex adjectives, and subordinate clauses) are pruned to shorten the sentence, the core informative entities (Noun Phrases) occupy a higher overall percentage of the remaining token count, resulting in a denser, fact-focused syntax.

\section{Conclusions}
\label{sec:orgb32ff03}

This study provides the first systematic comparison of semantic embedding spaces for sentence-level alignment in multilingual text simplification, a gap previously unaddressed in the literature.  We demonstrate that LaBSE's translation-ranking objective transfers robustly to monolingual paraphrase detection across Romance languages, while BGE-M3's retrieval-optimized architecture is slightly better for English. These findings are directly relevant to NLP researchers building simplification corpora, dataset curators working with comparable sources, and developers of accessibility tools targeting low-resource languages, all of whom require principled, reproducible methods for extracting high-precision parallel data from comparable sources without costly manual annotation.

\section{Limitations}
\label{sec:limitations}
While our pipeline successfully extracts high-precision parallel corpora, it has notable limitations. First, our strict thresholding strategy ($\tau$) prioritizes precision over recall, discarding approximately 95\% of the original sentences. Although this ensures a noise-free dataset, it inevitably filters out valid but highly abstract simplifications. Second, embedding performance is highly language-dependent; for instance, BGE-M3 excels in English but struggles with Romance languages. Finally, our framework currently operates exclusively at the sentence level, meaning fine-grained lexical substitutions or morphological adaptations are captured only implicitly within the aligned blocks. Extending this methodology to explicitly model token-level and sub-word alignments remains a primary focus of our ongoing work.

\section{Ethics Statement}
\label{sec:ethics}
This research complies with standard ethical guidelines for NLP. We utilize publicly available, crowdsourced data from Wikipedia and Vikidia, strictly adhering to their Creative Commons (CC-BY-SA) licenses. Given the encyclopedic nature of the texts, the dataset contains no personally identifiable information (PII) or sensitive personal data. By publicly releasing this multilingual sentence-aligned corpus, we aim to support the development of accessibility tools that assist children, language learners, and individuals with cognitive disabilities, thereby contributing to the democratization of information.

\section{Acknowledgments}
This document is part of a project that has received funding from the European Union’s Horizon Europe research and innovation program under Grant Agreement No. 101132431 (iDEM Project).  The University of Leeds was funded by UK Research and Innovation (UKRI) under the UK government’s Horizon Europe funding guarantee (Grant Agreement No. 10103529).  The views and opinions expressed in this document are solely those of the author(s) and do not necessarily reflect the views of the European Union. Neither the European Union nor the granting authority can be held responsible for them.

\section{Bibliographical References}\label{sec:reference}

\bibliographystyle{lrec2026-natbib}
\bibliography{bibexport}
\end{document}